\pdfoutput=1

\documentclass[11pt]{article}

\usepackage{enumitem}
\usepackage{amsmath}
\usepackage{tabularx}
\usepackage{makecell}
\usepackage{multirow}
\usepackage{graphicx}
\usepackage{float}

\usepackage[]{acl}

\usepackage{times}
\usepackage{latexsym}

\usepackage[T1]{fontenc}

\usepackage[utf8]{inputenc}

\usepackage{microtype}

\title{Mining both Commonality and Specificity from Multiple Documents for Multi-Document Summarization}

\author{Bing Ma \\
  Institute of Computing Technology, Chinese Academy of Sciences \\
  University of Chinese Academy of Sciences \\
  \texttt{mabingg28@gmail.com} \\}

\begin{document}
\maketitle
\begin{abstract}
The multi-document summarization task requires the designed summarizer to generate a short text that covers the important information of original documents and satisfies content diversity.
This paper proposes a multi-document summarization approach based on hierarchical clustering of documents. It utilizes the constructed class tree of documents to extract both the sentences reflecting the commonality of all documents and the sentences reflecting the specificity of some subclasses of these documents for generating a summary, so as to satisfy the coverage and diversity requirements of multi-document summarization. 
Comparative experiments with different variant approaches on \emph{DUC'2002-2004} datasets prove the effectiveness of mining both the commonality and specificity of documents for multi-document summarization. Experiments on \emph{DUC'2004} and \emph{Multi-News} datasets show that our approach achieves competitive performance compared to the state-of-the-art unsupervised and supervised approaches.
\end{abstract}

\section{Introduction}

Automatic text summarization is becoming much more important because of the exponential growth of digital textual information on the web.
Multi-document summarization, which aims to generate a short text containing important and diverse information of original multiple documents, is a challenging focus of NLP research. 
A well-organized summary of multiple documents needs to cover the main information of all documents comprehensively and simultaneously satisfy content diversity.
Extractive summarization approaches, which generate a summary by selecting a few important sentences from original documents, attract much attention because of its simplicity and robustness.
This paper focuses on extractive multi-document summarization.

Most extractive multi-document summarization approaches splice all the sentences contained in the original documents into a larger text, and then generate a summary by selecting sentences from the larger text \citep{lamsiyah2021unsupervised, yang2014enhancing, erkan2004lexrank}. 
However, the task of summarizing multiple documents is more difficult than the task of summarizing a single document.
Simply transforming multi-document summarization task into summarizing a single larger text completely breaks the constraints of documents on their sentences and lacks comparisons between documents, which results in the inability to mine the relevant information between documents, including mining the common information (commonality) of all documents and the important specific information (specificity) of some subclasses of documents.

The centroid-based summarization approaches focus on the commonality of all documents or all sentences and they select sentences based on the centroid words of all documents \citep{radev2004centroid, rossiello2017centroid} or the centroid embedding of all sentences \citep{lamsiyah2021unsupervised}. 
The clustering-based summarization approaches divide sentences into multiple groups and select sentences from each group \citep{yang2014enhancing, sarkar2009sentence}.
These approaches do not take into account the commonality and specificity of documents simultaneously.

Think about the process of human summarizing multiple documents: we would first describe the common information of all documents and then the important specific information of some subclasses of these documents respectively to satisfy the coverage and diversity requirements of multi-document summarization.

In this paper, inspired by the idea of human summarizing multiple documents, we propose a multi-document summarization approach based on hierarchical clustering of documents. 
Firstly, our model hierarchically clusters documents from top to bottom to build a class tree of documents.
Next, our model traverses each node along the class tree from top to bottom, and selects sentences from each node according to the similarity of sentences to the centroid embedding of the documents in the node and the dissimilarity to the centroid embedding of the documents not in the node, until the total length of the selected sentences reaches a pre-specified value. The sentence selected from the root node (containing all documents) reflects the commonality of all documents, and the sentence selected from each sub node (subclass) reflects the specificity of the subclass.
Finally, all selected sentences are arranged according to the order of their corresponding nodes on the class tree to form a summary.

Experiments are performed on standard datasets, including the \emph{DUC} datasets and the \emph{Multi-News} dataset. 
Comparative experiments on \emph{DUC'2002-2004} datasets prove that our approach considering both commonality and specificity of documents significantly outperforms the approaches considering only commonality or only specificity; And our approach (based on documents hierarchical clustering) outperforms the comparison approach based on sentences hierarchical clustering;
Experiments on \emph{DUC'2004} and \emph{Multi-News} datasets show that our approach outperforms lots of competitive supervised and unsupervised multi-document summarization approaches and yields considerable results.

In conclusion, our approach is unsupervised and easy to implement, and can be used as a strong baseline for evaluating multi-document summarization systems.

\section{Related Work}

The related works include centroid-based and clustering-based summarization methods.

The centroid-based methods score each sentence in documents by calculating the similarity between the sentence and the centroid of all documents or all sentences, so as to identify the most central sentences to generate a summary \citep{radev2004centroid, rossiello2017centroid, lamsiyah2021unsupervised}. The centroid-based methods focus on the commonality property of all documents or all sentences.
For example, MEAD \citep{radev2004centroid} scores each sentence based on the centroid words it contains (the words statistically important to multiple documents) and two other metrics (positional value and first-sentence overlap).
\citet{rossiello2017centroid} improves the original MEAD method by exploiting the word embedding representations to represent the centroid and each sentence and scoring each sentence based on the cosine similarity between the sentence embedding and the centroid embedding.
\citet{lamsiyah2021unsupervised} exploits sentence embedding model to represent each sentence and the centroid, and scores each sentence based on the cosine similarity between the sentence embedding and the centroid embedding, and two other metrics (sentence novelty and sentence position).

Many clustering-based extractive summarization methods cluster all sentences in documents and then select sentences from each sentence cluster to form a summary \citep{wang2008multi, mohd2020text, rouane2019combine, yang2014enhancing}.
For example, \citet{wang2008multi} groups sentences into clusters by sentence-level semantic analysis and symmetric non-negative matrix factorization, and selects the most informative sentences from each sentence cluster.
\citet{mohd2020text} represents each sentence as a big-vector using the Word2Vec model and applies the k-means algorithm to cluster sentences, and then scores sentences in each sentence cluster based on various statistical features (i.g. sentence length, position, etc.).
\citet{rouane2019combine} also uses k-means to cluster sentences, and then scores each sentence in each cluster based on the frequent itemsets of the cluster contained by the sentence.
\citet{yang2014enhancing} proposes a ranking-based sentence clustering framework to generate sentence clusters, and uses a modified MMR-like approach to select highest scored sentences from the descending order ranked sentence clusters to form the summary.

\section{Methodology}

Our approach takes a set of documents and a pre-given summary length as input, and outputs a multi-document summary.
It consists of three steps: (1) pre-processing of documents, (2) hierarchical clustering of documents for constructing a class tree of documents, and (3) sentence selection from the constructed class tree and summary generation.

\subsection{Pre-processing}

Pre-trained models are widely used in Natural Language Processing tasks. There are usually two ways to use the pre-trained models: (1) Feature Extraction based approach, which uses the pre-trained model learned from a large amount of textual data to encode texts of arbitrary length into vectors of fixed length; (2) Fine-Tuning based approach, which trains the downstream tasks by fine-tuning the pre-trained model's parameters.
In this paper, we adopt the feature extraction based approach, where the pre-trained model is applied on the input documents to obtain the embedding representations of sentences and documents.

Our model uses pre-trained sentence embeddings model to encode sentences.
And we can use two ways to obtain document embeddings: one is to directly obtain the document embeddings by taking each document as the input of the pre-trained embedding model; the other is to obtain the document embeddings based on sentence embeddings, e.g., a document embedding can be represented as the average of the sentence embeddings of all sentences it contains. 

Many pre-trained embedding models can be used in our model to obtain the sentence embeddings and document embeddings.
Here, we focus on the introduction of the proposed multi-document summarization approach, and the selection of pre-trained embedding models and document embedding methods are discussed later.

Formally, given a set of documents \emph{D} containing \emph{n} documents $ D=\{d_1,d_2,\cdots,d_n\} $.
Firstly, our model splits each document $ d_i \in D $ into sentences (denoted as $ d_i=\{s^i_1,s^i_2,\cdots,s^i_{|d_i|}\} $) using the Natural Language Toolkit (NLTK).\footnote{\href{https://www.nltk.org/api/nltk.tokenize.html}{nltk.tokenize}}
Next, our model maps each sentence in each document ($ s^i_k \in d_i $) to a fixed-length vector (denoted as $ \overrightarrow{s^i_k} $) using the pre-trained embedding model, and maps each document ($ d_i \in D $) to a vector of the same length (denoted as $ \overrightarrow{d_i} $).

\subsection{Hierarchical Clustering of Documents for Constructing Class Tree}

The proposed top-down hierarchical clustering algorithm for constructing the class tree of documents includes the following steps:

\textbf{Step 1: Generate the root node of class tree.} 

All documents in \emph{D} form the root node. The root node constitutes the first layer of class tree.

\textbf{Step 2: Generate the next layer of class tree.}

For each node of the latest layer of class tree, our model uses the k-means algorithm\footnote{\href{https://scikit-learn.org/stable/modules/generated/sklearn.cluster.KMeans.html}{sklearn.cluster.KMeans}} to divide the documents in the node into \emph{k} subnodes (also called \emph{k} subclasses). All new subnodes generated in this step constitute the new layer of class tree.

\textbf{Step 3: Repeat \emph{Step 2} until one of the following conditions is satisfied.}

\textbf{\emph{Condition 1}}: There is no node in the latest layer of class tree can be divided using the k-means algorithm.

\textbf{\emph{Condition 2}}: The total number of nodes on class tree exceeds the number of sentences required for the summary (specified or estimated according to the pre-given summary length). 

Because our model selects sentences from each node of class tree top-down until the pre-given value is reached.

\subsection{Generation of Summary}

After the construction of the class tree, our model traverses the nodes on the class tree from top to bottom and selects sentences from each node to generate a summary until the summary length reaches the pre-given length.

\emph{Section 3.3.1} introduces the overall flow of traversing the nodes on class tree to select sentences, \emph{Section 3.3.2} introduces the details of sentences scoring and selection in each node, and \emph{Section 3.3.3} introduces the process of sorting the selected sentences to form a summary.

\subsubsection{Overall Flow of Traversing Class Tree}

\begin{figure}
    \centering
    \includegraphics[scale=0.35]{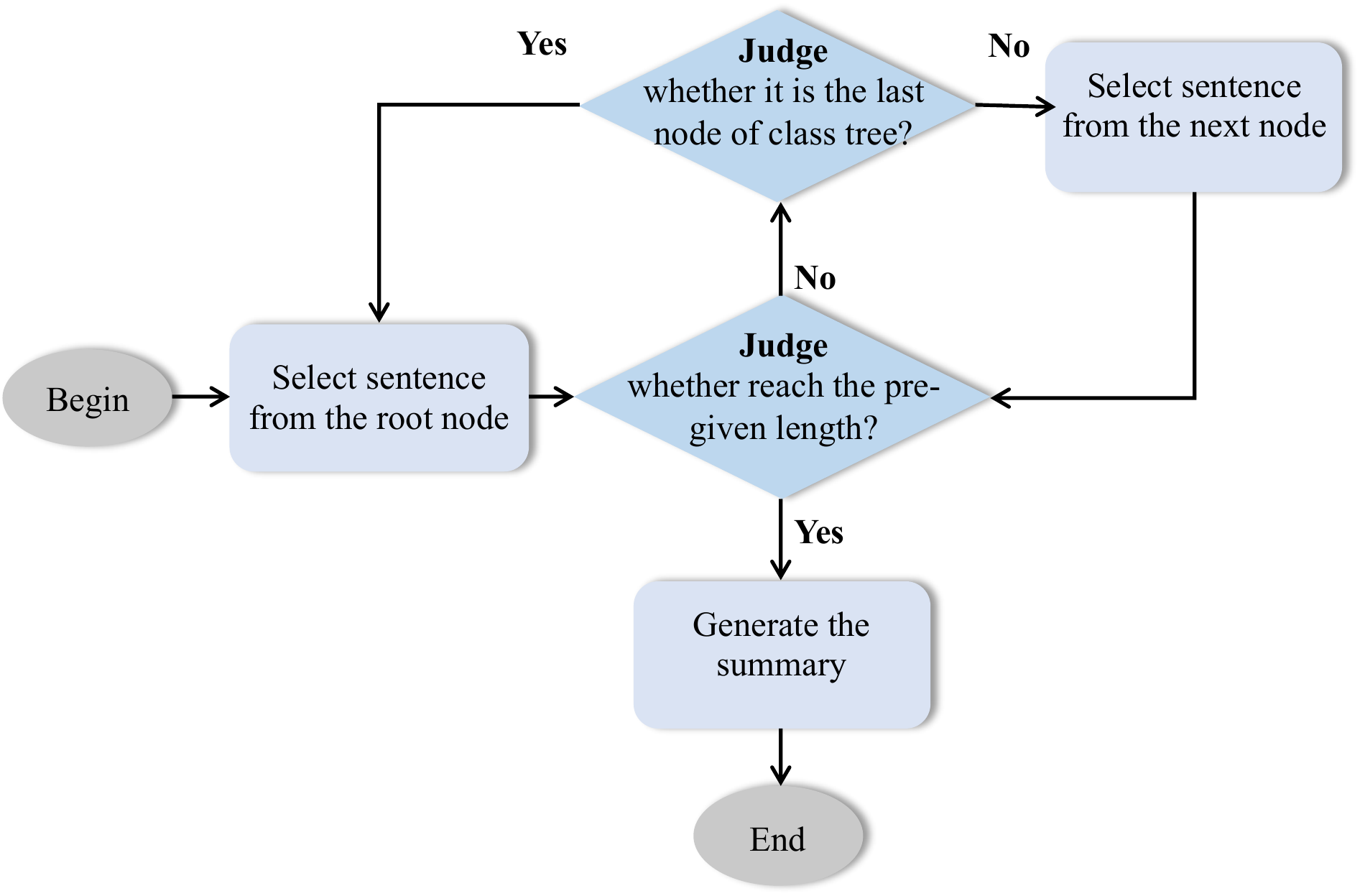}
    \caption{The flow chart of selecting sentences from the class tree.}
    \label{fig:Fig1}
\vspace{-1.3em}
\end{figure}

Fig.~\ref{fig:Fig1} displays the overall flow chart of traversing the nodes on the class tree for sentences selection.

The order of traversing the nodes on class tree follows two principles: (1) For different layers of class tree, our model traverses the layers from top to bottom; (2) For the nodes on the same layer, our model traverses the nodes in descending order by the number of documents contained in the nodes.
Because under the limitation of the pre-given summary length, our model hopes that the selected sentences can cover as many documents as possible while increasing diversity.

As shown in Fig.~\ref{fig:Fig1}, if the total length of the selected sentences does not reach the pre-given summary length after selecting sentence from the last node on the last layer of class tree, our model goes back to the first layer of class tree (the root node) to start the next iteration of selecting sentences, until the total length of all selected sentences in all iterations reaches the pre-given summary length.

\subsubsection{Sentence Scoring and Selection in Each Node}

Each node $ N_t $ on the class tree consists of multiple documents, denoted as $ N_t=\{d^t_1,\cdots,d^t_{|N_t|}\} $. Each document $ d^t_i \in N_t $ consists of multiple sentences, denoted as $ d^t_i=\{s^i_1,s^i_2,\cdots,s^i_{|d^t_i|}\} $.
Our model calculates the score of each sentence $ s^i_k $ in each node $ N_t $ ($ s^i_k \in d^t_i $ and $ d^t_i \in N_t $).

\noindent\textbf{(1) The Commonality-Specificity score of each sentence.} 

The centroid of all documents in $ N_t $ represents the common core of these documents. It is reasonable to think that the sentences that are more similar to the centroid of $ N_t $ are more relevant to the documents in $ N_t $, and the sentences that are less similar to the centroid of $ N_t $ are less relevant to the documents in $ N_t $.
Therefore, the Commonality-Specificity score of each sentence $ s^i_k $ in $ N_t $ can be calculated as the combination of its similarity to the centroid of $ N_t $ and its dissimilarity to the centroid of the documents not in $ N_t $.

Our model builds the centroid embedding vector of $ N_t $ (denoted as $ \overrightarrow{C_{N_t}} $) as the average of all document embedding vectors in it (as shown in Eq.~(\ref{eq:second})).
\begin{equation}
    \overrightarrow{C_{N_t}}=\frac{1}{|N_t|}\sum_{i=1}^{|N_t|} \overrightarrow{d^t_i} \label{eq:second}
\end{equation}
Where $ |N_t| $ denotes the number of documents in $ N_t $, and $ \overrightarrow{d^t_i} $ is the document embedding vector of the $i^{th}$ document in $ N_t $.

Similarly, the centroid embedding vector of the documents not in $ N_t $ (denoted as $ \overrightarrow{C_{\overline{N_t}}} $) is built as the average of all document embedding vectors not in $ N_t $ (i.e., $ \overline{N_t}=\{d \, | \, d \in D \, and \, d \notin N_t \} $).

The Commonality-Specificity score of each sentence $ s^i_k $ in $ N_t $ is calculated as follows:

\begin{equation}
\begin{aligned}
score^{CS}(s^i_k,N_t)=
\delta&* sim(\overrightarrow{s^i_k},\overrightarrow{C_{N_t}})\\
+(1\!-\!\delta)\!&*\! (1\!-\!sim(\overrightarrow{s^i_k},\overrightarrow{C_{\overline{N_t}}}))
\end{aligned}
\end{equation}
The value of $\delta \in [0,1]$. The larger value of $\delta$ illustrates more attention to the relevance with the documents in $N_t$, and the smaller value of $\delta$ illustrates more attention to the irrelevance with the documents not in $N_t$. When $\delta=1$, the $ score^{CS} $ of each sentence in $N_t$ only focuses on the relevance with the documents in $N_t$.
Our model uses the cosine similarity\footnote{\href{https://scikit-learn.org/stable/modules/generated/sklearn.metrics.pairwise.cosine_similarity.html}{sklearn.cosine\_similarity}} (denoted as $sim$) to calculate the similarity between embedding vectors.

The value of $ score^{CS} $ is bounded in $[0,1]$, and sentences with higher $ score^{CS} $ are considered to be more relevant to the documents in $ N_k $ and more irrelevant to the documents not in $N_k$.

\noindent\textbf{(2) Sentences scoring and selection.} 

We can use the Commonality-Specificity score alone to score and select sentences, or combine the Commonality-Specificity score with other scoring metrics to score and select sentences.

\noindent\textbf{\emph{Non-redundant score}}.
To reduce the redundancy of summary, our model would assign lower Non-redundant scores to the sentences that are more similar to the sentences already selected in previous steps.
Specifically, we use $S^p$ to represent the collection of sentences already selected in previous steps, the Non-redundant score of each sentence $ s^i_k $ in $ N_t $ is calculated as the dissimilarity between $ s^i_k $ and the sentence most similar to $ s^i_k $ in $S^p$, which is described as follows:
\begin{equation}
\begin{aligned}
    score^{NR}(s^i_k,N_t)=
    1\!-\!\max \limits_{s_p \in S^p}(sim(\overrightarrow{s^i_k},\overrightarrow{s_p}))
\end{aligned}
\end{equation}
The $score^{NR}$ is bounded in $[0,1]$, and sentences with higher $score^{NR}$ are considered to be less redundant with the sentences already selected in previous steps.
If $ S^p $ is Null (i.e., selecting the first sentence from the root node), the $score^{NR}$ of each sentence in the node is $1$.

\noindent\textbf{\emph{Position score}}.
Sentence position is one of the most effective heuristics for selecting sentences to generate summaries, especially for news articles \citep{edmundson1969new, ouyang-etal-2010-study}. We adopt the sentence position relevance metric (as Eq.~(\ref{eq:five})) introduced by \citet{joshi2019summcoder} to calculate the Position score of each sentence in each document.
\begin{equation}\label{eq:five}
\begin{aligned}
    score^P(s^i_k)=\max(0.5,\exp(\frac{-\mathcal{P}(s^i_k)}{\sqrt[3]{|d_i|}}))
\end{aligned}
\end{equation}
$\mathcal{P}(s^i_k)$ denotes the relative position of the $k^{th}$ sentence $s^i_k$ in the document $d_i$ (starting by $1$). 
The $score^P$ is bounded in $[0.5,1]$. The first sentence in each document obtains the highest $score^P$. The $score^P$ of sentences decrease as their distances from the beginning of documents increase, and remain stable at a value of $0.5$ after several sentences.

The final score of each sentence $s^i_k$ in $N_t$ can be defined as a linear combination of the three scores (as Eq.~(\ref{eq:six})).
\begin{equation}\label{eq:six}
\begin{aligned}
    &score^{final}(s^i_k,N_t)\!=\!\alpha\!*\!score^{CS}(s^i_k,N_t)\\
    +\!\beta&\!*\!score^{NR}(s^i_k,N_t)\!+\!\gamma\!*\!score^P(s^i_k)
\end{aligned}
\end{equation}
Where $\alpha+\beta+\gamma=1$, and $\alpha,\beta,\gamma\in[0,1]$. Different values of $\alpha$, $\beta$ and $\gamma$ represent different emphases on different scoring metrics.
Setting $\alpha=1$ means that we only use the Commonality-Specificity score to score sentences.

Our model selects the sentences with the highest final score that have not been selected in previous steps from $N_t$.
Only one sentence is selected from each node in each iteration because our model wants to traverse as many nodes as possible under the limitation of the pre-given summary length to increase the diversity of the generated summaries.

\subsubsection{Summary Generation}

After the process of selecting sentences from each node on the class tree, our model sorts the selected sentences to form a summary: (1) For the sentences selected from different nodes, our model sorts these sentences according to the traversal order of the nodes on the class tree (following the two principles introduced in \emph{Section 3.3.1}); (2) For multiple sentences selected from the same node (i.e., the first iteration of traversing the class tree for sentences selection does not select enough sentences), our model sorts the sentences according to the order in which they are selected.

The sentences selected from the root node express the commonality of all documents, and the sentences selected from each subnode express the specificity of the corresponding subclass.
The way of sorting sentences forms a summary with a total-sub structure.

\section{Experiment}

\subsection{Datasets and Evaluation Metrics}

\begin{table*}
\centering
\begin{tabular}{ccccc}
\hline
\textbf{Dataset} & \makecell[c]{\textbf{Number of} \\ \textbf{News Sets}} & \makecell[c]{\textbf{Number of} \\ \textbf{Docs}} & \makecell[c]{\textbf{Number of} \\ \textbf{Reference Summary} \\ of each news set} & \makecell[c]{\textbf{Number of Words} \\ of each sentence in Docs} \\
\hline
DUC'2002 & 59 & 567 & 2 & 22.86 \\
DUC'2003 & 30 & 298 & 4 & 25.43 \\
DUC'2004 & 50 & 500 & 4 & 25.38 \\
Multi-News & 5622 & 15326 & 1 & 22.24 \\
\hline
\end{tabular}
\caption{\label{data-description}
Description of each dataset, including the number of news sets, the total number of documents (news), the number of reference summaries for each news set and the average length of sentences in source documents.
}
\vspace{-1.3em}
\end{table*}

We evaluate the proposed approach on the standard multi-document summarization datasets, including \emph{DUC'2002-2004} datasets\footnote{\href{https://duc.nist.gov/}{DUC datasets}} and the \emph{Multi-News} dataset\footnote{\href{https://github.com/Alex-Fabbri/Multi-News}{Multi-News dataset}}. Table~\ref{data-description} describes the details of these datasets.
Each news set of \emph{DUC} datasets contains approximately $10$ documents on the same topic.
Each news set of \emph{Multi-News} dataset contains a different number of documents (from $1$ to $10$) on the same topic \citep{fabbri-etal-2019-multi}.
Each news set has several reference summaries written by experts.

ROUGE is a standard evaluation metric for automatic document summarization \citep{lin-2004-rouge}, including \emph{ROUGE-1}, \emph{ROUGE-2}, \emph{ROUGE-L} and \emph{ROUGE-SU4} (denoted as \emph{R-1}, \emph{R-2}, \emph{R-L} and \emph{R-SU4} respectively).
We use the ROUGE toolkit (version \emph{1.5.5}), and adopt the same ROUGE settings\footnote{ROUGE-1.5.5 with parameters "-n 2 -2 4 -u -m -r 1000 -f A -p 0.5" and "-l 100" for DUC'2002 and DUC'2003; "-b 665" for DUC'2004; "-l 264" for Multi-News.} that are commonly used on the \emph{DUC} datasets and \emph{Multi-News} dataset for multi-document summarization.
Guided by the state-of-the-art approaches, we report \emph{ROUGE recall} on \emph{DUC} datasets and \emph{ROUGE F1-score} on the \emph{Multi-News} dataset, respectively.

\subsection{Experimental Settings}

\noindent\textbf{(1) Selection of Pre-trained Model} 

\citet{lamsiyah2021unsupervised} has studied multi-document summarization based on centroid approach and sentence embeddings, which verifies the effectiveness of sentence embedding representations for multi-document summarization, and shows that using different sentence embedding models would affect the performance of summarization. 
Its experiment results show that the \emph{USE-DAN} model \citep{cer2018universal} is one of the best performing models.

In order to focus on the verification of the effectiveness of our proposed multi-document summarization approach and not affected by different embedding models, we use the \emph{USE-DAN} model\footnote{\href{https://tfhub.dev/google/universal-sentence-encoder/4}{universal-sentence-encoder}} to encode sentences.
In order to unify the expression of sentences and documents and preserve the relationship between documents and sentences, we obtain the embedding vector of each document $ d_i \in D $ by calculating the average of the sentence embedding vectors of all sentences it contains.

The comparison of using different pre-trained embedding models and different document embedding methods in our proposed summarization approach will be discussed in the follow-up work, such as the pre-trained models GPT-3, Bert, and so on.

\noindent\textbf{(2) Determination of Hyperparameters} 

Different values of hyperparameters affect the results of the proposed approach, and we determined their values both theoretically and experimentally.

\textbf{Estimation of the hyperparameter \emph{k} in k-means algorithm.}
Our model needs to select sentences not only from the root node of the class tree, but also from as many sub nodes as possible, to extract both commonality and specificity information of the input documents.
Thus, when generating the sub nodes of the second layer of class tree, the value of \emph{k} in k-means clustering should not be set too large. Otherwise, under the limitation of the pre-given summary length, the sub nodes participating in sentences selection cannot cover all input documents, resulting in the generated summary cannot contain the specificity information of some subclasses of the input documents.

The approximate number of sentences need to be selected for generating summaries can be estimated by \emph{(average length of target summaries) $\div$ (average length of sentences in source documents)}, i.e., $4.65$ for \emph{DUC'2004}, $3.93$ for \emph{DUC'2003}, $4.37$ for \emph{DUC'2002}. 
Based on the estimation, when generating the sub nodes of the second layer, the hyperparameter \emph{k} of the k-means clustering is set to be within the range of $[2,4]$ (\emph{minimum number of sentences estimated for different datasets$-1$}). 
For simplicity, when generating the sub nodes of the third layer and subsequent layers, we set \emph{k} in k-means clustering to $2$.

\textbf{Estimation of the hyperparameters $\delta$ and $\alpha$, $\beta$, $\gamma$ in sentences scoring equation.}
The hyperparameter $\delta$ in $score^{CS}$ illustrates the concern for the relevance of each sentence to the documents in its own node, so theoretically it cannot be set too small. 
The hyperparameters $\alpha$, $\beta$ and $\gamma$ in $score^{final}$ illustrate different degrees of attention to the three scores. 
Theoretically $\alpha$ cannot be set too small because our approach focuses on extracting both commonality and specificity of documents for multi-document summarization.

In order to determine the exact values of these hyperparameters, we employed a procedure similar to that used by \citet{lamsiyah2021unsupervised} and \citet{joshi2019summcoder}. We built a small held-out set by randomly sampling 25 clusters with different length of reference summaries from the validation set of the \emph{Multi-News} dataset. Then we performed a grid search for these hyperparameters: $\delta$, $\alpha$, $\beta$, $\gamma$ $\in [0,1]$ with constant step of 0.1 under the condition $\alpha+\beta+\gamma=1$, $k\in[2,4]$ with constant step of 1.
Finally, the obtained values of the hyperparameters are $3$, $0.9$, $0.8$, $0.1$, $0.1$ for $k$, $\delta$, $\alpha$, $\beta$ and $\gamma$ respectively, which are consistent with the theoretical analysis of these hyperparameters.

\subsection{Evaluations}

We evaluate the proposed approach from three aspects: (1) verify the effectiveness of mining both commonality and specificity of documents for multi-document summarization; (2) verify the effectiveness of using documents hierarchical clustering for multi-document summarization compared to using sentences hierarchical clustering; (3) verify the effectiveness of the proposed multi-document summarization approach compared to other unsupervised and supervised approaches.
Due to the randomness of k-means, each experiment has been run three times to get the intermediate results.

\noindent\textbf{(1) Verify the effectiveness of mining both commonality and specificity of documents for multi-document summarization.}

In order not to be affected by other factors, in this part, our model uses only the \emph{Commonality-Specificity score} to score sentences, denoted as \emph{\textbf{Ours$\mathbf{_{CS}}$}} (i.e., $\alpha=1,\beta=0,\gamma=0$). 
Different variant approaches are designed as follows:

\noindent\emph{\textbf{Comp$\mathbf{_1}$}}: The variant approach \emph{Comp$\mathbf{_1}$} only focuses on the commonality of all documents. It does not use any clustering and only scores sentences by calculating the cosine similarity between sentence embeddings and the centroid embedding of all documents.

\noindent\emph{\textbf{Comp$\mathbf{_2}$}}: The variant approach \emph{Comp$\mathbf{_2}$} only focuses on the specificity of each subcluster of documents. It uses k-means algorithm to cluster documents, and then uses the \emph{Commonality-Specificity score} to score sentences in each subcluster and select sentences from each subcluster.

\noindent\emph{\textbf{Comp$\mathbf{_3}$}}: The variant approach \emph{Comp$\mathbf{_3}$} is similar to \emph{Comp$_2$} but \emph{Comp$\mathbf{_3}$} scores sentences in each subcluster by only calculating the cosine similarity between the sentence embeddings and the centroid embedding of the subcluster. i.e., for each subcluster, \emph{Comp$\mathbf{_3}$} only focuses on the relevance with the documents in the subcluster and ignores the irrelevance with the documents not in it.

\begin{table}
\centering
\setlength{\tabcolsep}{1.2mm}{
\begin{tabular}{cccccc}
\hline
 & \textbf{Method} & \textbf{R-1} & \textbf{R-2} & \textbf{R-L} & \textbf{R-SU4} \\
\hline
\multirow{4}{*}{DUC'2004} 
& \textbf{Ours}$\mathbf{_{CS}}$ & \textbf{38.36} & \textbf{8.66} & \textbf{33.94} & \textbf{13.30} \\
& Comp$_1$ & 36.44 & 8.03 & 32.08 & 12.61 \\
& Comp$_2$ & 36.38 & 7.91 & 31.78 & 12.28 \\
& Comp$_3$ & 36.85 & 8.18 & 32.25 & 12.44 \\
\hline
\multirow{4}{*}{DUC'2003} 
& \textbf{Ours}$\mathbf{_{CS}}$ & \textbf{37.89} & \textbf{8.27} & \textbf{32.37} & \textbf{12.85} \\
& Comp$_1$ & 36.77 & 8.22 & 31.29 & 12.7 \\
& Comp$_2$ & 35.31 & 7.05 & 30.24 & 11.39 \\
& Comp$_3$ & 35.03 & 7.02 & 30.07 & 11.29 \\
\hline
\multirow{4}{*}{DUC'2002} 
& \textbf{Ours}$\mathbf{_{CS}}$ & \textbf{34.81} & \textbf{7.32} & \textbf{30.11} & \textbf{11.48} \\
& Comp$_1$ & 33.23 & 6.95 & 28.78 & 10.99 \\
& Comp$_2$ & 32.90 & 6.20 & 28.37 & 10.37 \\
& Comp$_3$ & 32.86 & 6.12 & 28.46 & 10.31 \\
\hline
\end{tabular}
}
\caption{\label{group1}
Comparison results of different approaches regarding whether or not the commonality and specificity of documents are considered on \emph{DUC} datasets.
}
\vspace{-0.9em}
\end{table}

Table~\ref{group1} displays the results on three \emph{DUC} datasets.
By comparison with \emph{Comp$_1$}, \emph{Comp$_2$} and \emph{Comp$_3$}, our approach outperforms all these variant approaches on all metrics. Because our approach first selects sentence based on the commonality of all documents, and then selects sentences based on the specificity of different subclasses, which is in line with the way of human summarizing multiple documents. 

\noindent\textbf{(2) Verify the effectiveness of using documents hierarchical clustering for multi-document summarization.}

In order not to be affected by other factors, in this part, our model also uses only the \emph{Commonality-Specificity score} to score sentences (\emph{Ours$_{CS}$}). For a fair comparison, the comparison experiment is designed as follows:

\noindent\emph{\textbf{Comp$\mathbf{_4}$}}: The variant approach \emph{Comp$_{4}$} is similar to \emph{Ours$_{CS}$} but \emph{Comp$_{4}$} uses all sentences of the input documents as input and hierarchically clusters all sentences, and then selects sentences from the class tree of sentence clusters.

\begin{table}
\vspace{1.0em}
\centering
\setlength{\tabcolsep}{1.2mm}{
\begin{tabular}{cccccc}
\hline
 & \textbf{Method} & \textbf{R-1} & \textbf{R-2} & \textbf{R-L} & \textbf{R-SU4} \\
\hline
\multirow{2}{*}{DUC'2004} 
& \textbf{Ours}$\mathbf{_{CS}}$ & \textbf{38.36} & \textbf{8.66} & \textbf{33.94} & \textbf{13.30} \\
& Comp$_4$ & 37.17 & 7.65 & 32.93 & 12.45 \\
\hline
\multirow{2}{*}{DUC'2003} 
& \textbf{Ours}$\mathbf{_{CS}}$ & \textbf{37.89} & \textbf{8.27} & \textbf{32.37} & \textbf{12.85} \\
& Comp$_4$ & 36.63 & 7.88 & 31.48 & 12.31 \\
\hline
\multirow{2}{*}{DUC'2002} 
& \textbf{Ours}$\mathbf{_{CS}}$ & \textbf{34.81} & \textbf{7.32} & \textbf{30.11} & \textbf{11.48} \\
& Comp$_4$ & 33.50 & 6.48 & 29.09 & 10.63 \\
\hline
\end{tabular}
}
\caption{\label{group2}
Comparison results of documents hierarchical clustering-based approach and sentences hierarchical clustering-based approach on \emph{DUC} datasets.
}
\vspace{-1.3em}
\end{table}

Table~\ref{group2} displays the results on three \emph{DUC} datasets. 
By comparison with \emph{Comp$_{4}$}, our approach based on documents hierarchical clustering outperforms the variant approach that based on sentences hierarchical clustering on all metrics.
Because the sentences hierarchical clustering-based approach lacks comparisons between documents, thus resulting in the inability to discover the relationships between documents, which are important for multi-document summarization.

\noindent\textbf{(3) Compare the proposed approach with other multi-document summarization approaches.}

Table~\ref{group3} and Table~\ref{group4} compare the performance of our approach with existing state-of-the-art multi-document summarization approaches on \emph{DUC'2004} and \emph{Multi-News} datasets respectively.
\emph{\textbf{Ours$\mathbf{_{CS}}$}} corresponds to our approach that uses only the Commonality-Specificity score, and \emph{\textbf{Ours$\mathbf{_{final}}$}} corresponds to our approach that score sentences using $score^{final}$, i.e., the combination of the three scores.


\begin{table*}
\centering
\setlength{\tabcolsep}{1.3mm}{
\begin{tabular}{lcccc}
\hline
\textbf{Method} & \textbf{R-1} & \textbf{R-2} & \textbf{R-L} & \textbf{R-SU4} \\
\hline
\multicolumn{5}{l}{\emph{Unsupervised methods}} \\
Lead & 32.37 & 6.38 & 28.68 & 10.29 \\
LexRank\citep{erkan2004lexrank} & 37.32 & 7.84 & 33.18 & 12.53 \\
Centroid$_{BOW}$\citep{radev2004centroid} & 37.03 & 8.19 & 32.48 & 12.68 \\
GreedyKL\citep{hong2014repository} & 37.99 & 8.54 & 33.03 & 13.02 \\
OCCAMS\_V\citep{davis2012occams} & 37.51 & 9.42$^*$ & 33.69 & 13.12 \\
Ranking-clustering\citep{yang2014enhancing} & 37.87 & 9.35 & - & 13.25 \\
SummPip\citep{zhao2020summpip} & 36.30 & 8.47 & - & 11.55 \\
Centroid$^{Run1}_{embedding}$\citep{lamsiyah2021unsupervised} & 36.92 & 8.20 & 32.53 & 12.72 \\
Centroid$^{Run4}_{embedding}$\citep{lamsiyah2021unsupervised} & 38.12 & 9.07 & 34.15 & 13.44 \\
\hline
\multicolumn{5}{l}{\emph{Supervised methods}} \\
PG-MMR\citep{lebanoff-etal-2018-adapting} & 36.42 & 9.36 & - & 13.23 \\
CopyTransformer\citep{gehrmann-etal-2018-bottom} & 28.54 & 6.38 & - & 7.22 \\
Hi-MAP\citep{fabbri-etal-2019-multi} & 35.78 & 8.90 & - & 11.43 \\
BART-Long-Graph\citep{pasunuru-etal-2021-efficiently} & 34.72 & 7.97 & - & 11.04 \\
Primera\citep{xiao-etal-2022-primera} & 35.1 & 7.2 & 17.9 & - \\
\hline
$\mathbf{Ours_{final}}$ & \textbf{39.28$^*$} & \textbf{9.31} & \textbf{35.02$^*$} & \textbf{13.75$^*$} \\
$\mathbf{Ours_{CS}}$ & \textbf{38.36} & \textbf{8.66} & \textbf{33.94} & \textbf{13.30} \\
\hline
\end{tabular}
}
\caption{\label{group3}
ROUGE scores of different approaches on \emph{DUC'2004} dataset. The best performing approach for each metric is indicated by *.
}
\vspace{-1.3em}
\end{table*} 

We compare our approach with both unsupervised approaches and supervised deep learning-based approaches.
The unsupervised approaches listed are some competitive baselines or state-of-the-art approaches for extractive multi-document summarization.
We reproduce \emph{Centroid$_{embedding}$} \citep{lamsiyah2021unsupervised} using the \emph{USE-DAN} sentence embedding model and list its results on \emph{DUC'2004} to compare with our approach, because \emph{Centroid$_{embedding}$} and our approach use the same sentence embedding model but \emph{Centroid$_{embedding}$} only focuses on the commonality of all documents.
The results of other approaches are directly taken from their original articles \citep{hong2014repository} or published materials\footnote{\href{https://github.com/stuartmackie/duc-2004-rouge}{github/duc2004-results}}.
The supervised approaches are first trained on large datasets, such as \emph{CNN}, \emph{DailyMail} and \emph{Multi-News}, and then tested on \emph{DUC'2004} and \emph{Multi-News} datasets. The results are directly taken from their original articles.

As shown in Table~\ref{group3}, for \emph{R-1} measure, our approach, both Ours$_{CS}$ and Ours$_{final}$, significantly outperforms all unsupervised and supervised approaches. 
And for \emph{R-L} and \emph{R-SU4} measures, our approach Ours$_{final}$ significantly outperforms all listed approaches. 
For \emph{R-2} measure, our approach achieves comparable result with the state-of-the-art approaches. The supervised approaches yield worse results on \emph{DUC'2004} than most unsupervised approaches because these deep learning-based approaches are trained on other datasets and tested directly on \emph{DUC'2004}.
The comparison results with \emph{Centroid$_{embedding}$} further illustrate the effectiveness of mining both commonality and specificity for multi-document summarization.

As shown in Table~\ref{group4}, our approach significantly outperforms all unsupervised approaches on \emph{R-1}, \emph{R-L} and \emph{R-SU4} metrics. 
By comparing with the supervised deep learning-based approaches that are trained and tested on \emph{Multi-News} dataset, our approach still achieves significantly better \emph{R-1}, \emph{R-L} and \emph{R-SU4} scores than \emph{PG-MMR}, \emph{CopyTransformer}, \emph{Hi-MAP} and \emph{DynE} approaches.
For \emph{R-L} measure, our approach achieves best result than all listed approaches.

Overall, as an unsupervised and easy-to-implement approach, our model achieve considerable results.
Moreover, the comparison experiments with different variant approaches prove the effectiveness of mining both the commonality and specificity of documents for multi-document summarization.

\begin{table*}
\centering
\setlength{\tabcolsep}{1.0mm}{
\begin{tabular}{lcccc}
\hline
\textbf{Method} & \textbf{R-1} & \textbf{R-2} & \textbf{R-L} & \textbf{R-SU4} \\
\hline
\multicolumn{5}{l}{\emph{Unsupervised methods}} \\
Lead & 39.41 & 11.77 & - & 14.51 \\
MMR & 38.77 & 11.98 & - & 12.91 \\
LexRank\citep{erkan2004lexrank} & 38.27 & 12.70 & - & 13.20 \\
SummPip\citep{zhao2020summpip} & 42.32 & 13.28 & - & 16.20 \\
Spectral-BERT\citep{wang2020spectral} & 40.9 & 13.6 & - & 16.7 \\
BART$_{fine-tuned}$\citep{johner2021error} & 40.58 & 15.50 & 21.73 & - \\
Centroid$^{Run4}_{embedding}$\citep{lamsiyah2021unsupervised} & 42.93 & 14.04 & 27.7 & 17.27 \\
\hline
\multicolumn{5}{l}{\emph{Supervised methods}} \\
PG-MMR\citep{lebanoff-etal-2018-adapting} & 40.55 & 12.36 & - & 15.87 \\
CopyTransformer\citep{gehrmann-etal-2018-bottom} & 43.57 & 14.03 & - & 17.37 \\
Hi-MAP\citep{fabbri-etal-2019-multi} & 43.47 & 14.89 & - & 17.41 \\
DynE\citep{hokamp2020dyne} & 43.9 & 15.8$^*$ & 22.2 & - \\
MGSum\citep{jin-etal-2020-multi} & 44.75$^*$ & 15.75 & - & 19.30$^*$ \\
\hline
\textbf{Ours$\mathbf{_{final}}$} & \textbf{44.04} & \textbf{14.15} & \textbf{39.74$^*$} & \textbf{18.19} \\
\hline
\end{tabular}
}
\caption{\label{group4}
ROUGE scores of different approaches on \emph{Multi-News} dataset. The best performing approach for each metric is indicated by *.
}
\vspace{-1.3em}
\end{table*}

\section{Conclusion and Future Work}

In this paper, we propose a multi-document summarization approach based on hierarchical clustering of documents, which makes use of the generated class tree of documents to mine both the commonality of all documents and the important specificity of some subclasses of documents, so as to generate the summary in line with human summarizing multiple documents. In the experiments, we show that our approach significantly outperforms the variant approaches mining only commonality or only specificity, and the variant approach based on sentences hierarchical clustering. Furthermore, as an easy-to-implement unsupervised approach, our approach is superior to many competitive supervised and unsupervised multi-document summarization approaches, and yields considerable results.

Documents hierarchical clustering has been proven to be effective for multi-document summarization in this paper. In future work, we plan to explore other effective hierarchical clustering approaches for multi-document summarization or explore the best \emph{k} of k-means in hierarchical clustering. Additionally, we will compare different document embedding methods for hierarchical clustering and multi-document summarization to explore the suitable embedding representation of documents.

\bibliography{anthology,custom}
\bibliographystyle{acl_natbib}






\end{document}